\documentclass{article}

\usepackage{times}
\usepackage{graphicx} 
\usepackage{subfigure} 

\usepackage{natbib}

\usepackage{algorithm}
\usepackage{algorithmic}

\usepackage{hyperref}

\usepackage[accepted]{icml2015} 

\begin{document} 

\twocolumn[
\icmltitle{Feature Learning for Interaction Activity Recognition in RGBD Videos}

\icmlauthor{Ngu Nguyen}{nlnngu@fit.hcmus.edu.vn}
\icmladdress{University of Science,
            Ho Chi Minh City, Vietnam}

\icmlkeywords{boring formatting information, machine learning, ICML}

\vskip 0.3in
]

\begin{abstract} 
This paper proposes a human activity recognition method which is based on features learned from 3D video data without incorporating domain knowledge.
The experiments on data collected by RGBD cameras produce results outperforming other techniques.
Our feature encoding method follows the bag-of-visual-word model, then we use a SVM classifier to recognise the activities.
We do not use skeleton or tracking information and the same technique is applied on color and depth data.
\end{abstract} 

\section{Introduction}
\label{sec:intro}

Human activity recognition leverage various sensing technologies and provide a vast range of potential applications.
Researchers in computer vision have reached a large number of achievements in activity analysis \cite{2011_Ryoo_HumanActivityAnalysis}.
However, the vision-based approach suffers from issues related to obtrusiveness and complexity of real-world settings.
Low-cost RGBD cameras, such as Microsoft Kinect devices, provide both color and depth information, which can improve activity recognition in accuracy and robustness~\cite{2014_Aggarwal_3DReview}.

In this paper, we propose a method to extract features from RGBD videos.
The key component of our approach is a Independent Subspace Analysis network~\cite{2011_Le_ISA}  
Our approach does not rely on hand-crafted features, but learns discriminative features from the input data.
The same method is performed to extract features from all modalities.
Our proposed method directly exploit color (grayscale) and depth data to extract features for human activity recognition.
Moreover, there is no domain knowledge that is incorporated in this method.
That means it is possible to apply the technique for various applications.
Chen \textit{et al.}~\cite{2014_Chen_ISACombination} also utilized ISA networks but they worked only on depth data and relied on skeleton data to extract features.
Our method directly operates on color and depth data.
It is essential because the skeleton information is not always available, due to sensor noises or self-occlusions of human bodies, especially in collaborative activities involving two or more persons.

The rest of this paper is organized as follows.
In Section~\ref{sec:single}, we present the ISA-based feature learning method.
The performance of our approach is evaluated in Section~\ref{sec:results} through experimental results on interaction activities between two persons.
Finally, we conclude the paper and introduce possible extension in Section~\ref{sec:conclusion}.

\section{Feature Learning using Independent Subspace Analysis}
\label{sec:single}

Independent Subspace Analysis (ISA), an extension of Independent Component Analysis (ICA), is widely-used in the field of natural image statistics~\cite{2009_Hyvrinen_NaturalImgStat}.
ISA can extract features that are invariant to local translation and selective to frequency, rotation and velocity~\cite{2011_Le_ISA}.
Le \textit{et al.}~\cite{2011_Le_ISA} adapted the technique to recognize activities in videos.
Their method achieved state-of-the-art at that time on well-known benchmark datasets. 
Multiple ISA layer can be stacked to form an ISA network, where outputs of one layer are inputs of the above layer.
Thus, the ISA network is possible to learn a hierarchical representation of the input data.
Bag-of-word model can be applied to form feature vectors for a classification algorithm.

\begin{figure}[!htb]
	\centering
	\includegraphics[width=8cm]{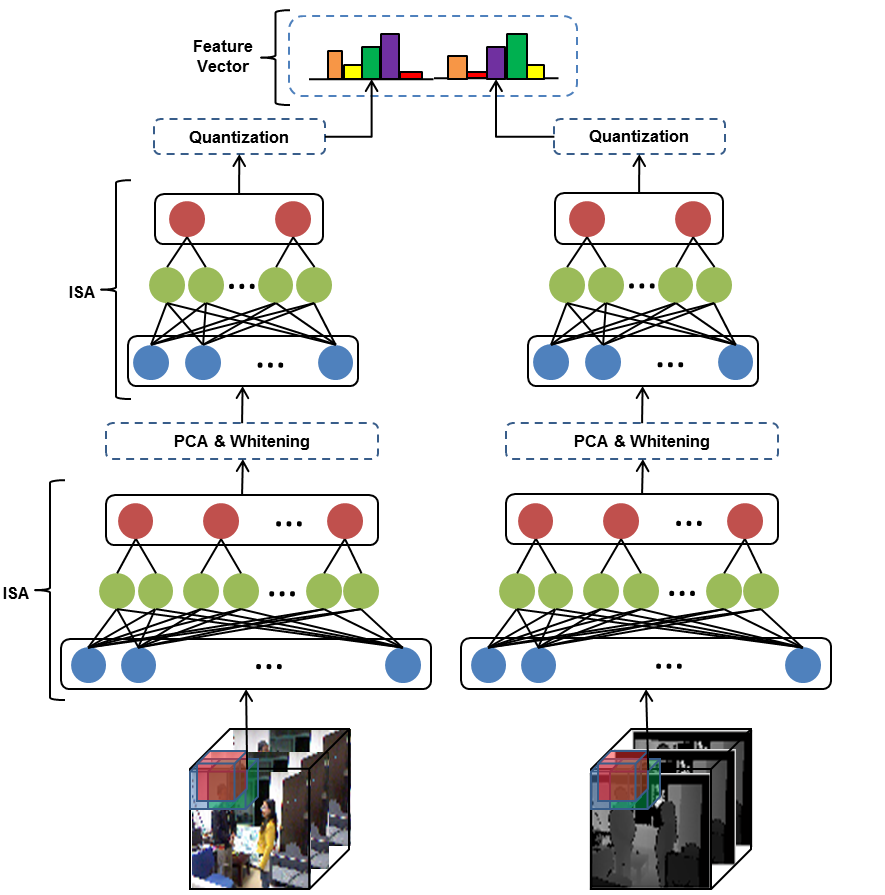}
	\caption{Our proposed framework, which learns discriminative features from color and depth data}
	\label{fig:featurelearning}
\end{figure}

Using the method proposed by Le \textit{et al.}~\cite{2011_Le_ISA}, we extract the features from unlabeled data of each single modality, e.g grayscale and depth video.
In this paper, we process grayscale and depth data in the same way (Figure~\ref{fig:featurelearning}).
We first take a spatial-temporal data block from each modality and flatten it frame-by-frame into a vector.
Then, this vector is the input of the ISA network.
To train the ISA network, we use batch projected gradient descent, which is the same technique in~\cite{2011_Le_ISA}.
Finally, we use the \textit{pre-trained} network and k-means algorithm to generate the histogram-based feature vectors for a SVM-based classifier.

\section{Experimental Results}
\label{sec:results}

We performed experiments on the second subset OA2 of the Office Activity dataset~\cite{2014_Wang_ConvolutionalNetworks}, which contains ten interaction activities: \textit{Asking-and-away}, \textit{Called-away}, \textit{Carrying}, \textit{Chatting}, \textit{Delivering}, \textit{Eating-and-chatting}, \textit{Having-guest}, \textit{Showing}, \textit{Seeking-help}, \textit{Shaking-hands}.
Ten subjects performed these activities in two different offices.
The authors divided the dataset according to the name of one person in each interaction session.

Our experiments are implemented on a desktop computer with Intel Core i7 CPU and 16 GB RAM.
We first \textit{pre-train} the ISA network with unlabeled data.
Then, we extract features using the network and form the bag-of-visual-word representation for each video clip.
Finally, a SVM-based classifier is used to recognize the activities.
We used the RBF kernel and performed \textit{grid search} to select the optimal parameters.
We performed experiments on the OA2 dataset~\cite{2014_Wang_ConvolutionalNetworks} with image size $80\times60$.
The features extracted by our two-layer ISA network are distributed into 100 clusters, from which we build the bag-of-word model.
Thus, each videos are represented with a 100-dimension vector.
We after that follow the same cross-validation scheme as described in~\cite{2014_Wang_ConvolutionalNetworks} (i.e \textit{leave-one-person-out}).

\begin{figure}[!htb]
	\centering
	\includegraphics[width=8cm]{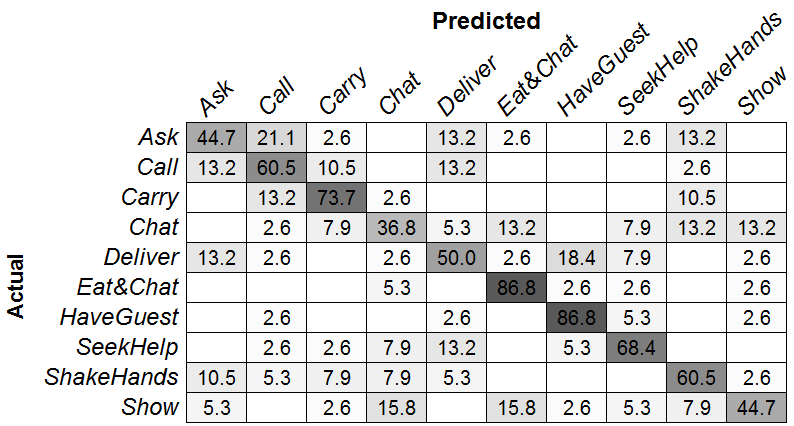}
	\caption{Confusion matrix on OA2 dataset~\cite{2014_Wang_ConvolutionalNetworks}}
	\label{fig:confusionmatrixoa2}
\end{figure}

Table \ref{tab:OA2dataset} and Table~\ref{tab:OA2dataset2} show the average accuracy of \textit{leave-one-person-out} validation scheme on the OA2 dataset, comparing with state-of-the-art results provided in the paper of Wang \textit{et al.}~\cite{2014_Wang_ConvolutionalNetworks}.
Most confusion is due to the semantic meaning of activities (Figure~\ref{fig:confusionmatrixoa2}).
For instance, \textit{Ask} is very similar with \textit{Call} in real life.
To solve this issue, we may integrate new modalities into the system, e.g microphone for audio recording.

\begin{table}[t]
	\begin{center}
	\caption{Average accuracy on the OA2 dataset~ \cite{2014_Wang_ConvolutionalNetworks}}
	\label{tab:OA2dataset}
	\begin{tabular}{|c|c|c|}
  		\hline
  		\textbf{} & \cite{2014_Wang_ConvolutionalNetworks} & Ours
  		\\
  		\hline
  		\textit{Grayscale} & 41.6\% & 60.0\% \\
  		\hline
  		\textit{Depth} & 43.6\% & 50.8\% \\
  		\hline
  		\textit{Grayscale \& Depth} & 45.0\% & 61.3\% \\
  		\hline
\end{tabular}
\end{center}
\end{table}

\begin{table}[t]
\begin{center}
\caption{Comparison of the accuracy per activity category on the OA2 dataset with the results in~\cite{2014_Wang_ConvolutionalNetworks}. DCSF, ConvNet, and ConfNet are the methods of \cite{2013_Xia_DepthCuboid}, \cite{2013_Ji_3DNeuralNet}, and \cite{2014_Wang_ConvolutionalNetworks}, respectively.}
	\label{tab:OA2dataset2}
\begin{tabular}{|l|c|c|c|c|}
\hline
                    & DCSF       & ConvNet & ConfNet & Ours             \\ \hline
\textit{Ask}        & 12.5\%          & 39.6\%   & 25.3\%     & \textbf{44.7\%} \\ \hline
\textit{Call}       & 45.8\%          & 44.8\%   & 57.5\%     & \textbf{60.5\%} \\ \hline
\textit{Carry}      & 66.7\%          & 56.8\%   & 53.5\%     & \textbf{73.7\%} \\ \hline
\textit{Chat}       & \textbf{37.5\%} & 17.2\%   & 25.3\%     & 36.8\%          \\ \hline
\textit{Deliver}    & 20.1\%          & 34.5\%   & 32.8\%     & \textbf{50.0\%}    \\ \hline
\textit{Eat\&Chat}  & 50.0\%          & 35.8\%   & 69.5\%     & \textbf{86.8\%} \\ \hline
\textit{HaveGuest}  & 37.5\%          & 34.1\%   & 43.7\%     & \textbf{86.8\%} \\ \hline
\textit{SeekHelp}   & 16.7\%          & 44.8\%   & 59.2\%     & \textbf{68.4\%} \\ \hline
\textit{ShakeHands} & 41.7\%          & 32.8\%   & 59.8\%     & \textbf{60.5\%} \\ \hline
\textit{Show}       & 37.5\%          & 29.3\%   & 23.0\%     & \textbf{44.7\%} \\ \hline
\end{tabular}
\end{center}
\end{table}

\section{Conclusion}
\label{sec:conclusion}
In this paper, we introduce a unsupervised learning method that extracts features from color and depth data.
The proposed technique is used to recognize human activities in RGBD videos.
The features are extracted from raw color and depth data, without relying on skeleton or tracking information.
Our experimental results show that using learned features helps to improve the classification accuracy.
Moreover, the approach is generic enough to apply in other applications.
In future, more sensing modalities (e.g. microphone) can be integrated to improve the accuracy of our proposed approach.

\bibliography{example_paper}
\bibliographystyle{icml2015}

\end{document}